\documentclass[sigconf]{acmart}
\usepackage{makecell}
\usepackage{multirow}
\usepackage{bigstrut}
\usepackage[inkscapelatex=false]{svg}
\usepackage[justification=centering]{caption}
\usepackage{color}
\usepackage{tcolorbox}
\usepackage{CJK}
\usepackage{adjustbox}
\usepackage{threeparttable}
\tcbset{width=1\textwidth,boxrule=0pt,colback=red,arc=0pt,auto outer arc,left=0pt,right=0pt,boxsep=1pt}
\AtBeginDocument{%
  \providecommand\BibTeX{{%
    \normalfont B\kern-0.5em{\scshape i\kern-0.25em b}\kern-0.8em\TeX}}}

\copyrightyear{2023}
\acmYear{2023}
\setcopyright{acmlicensed}\acmConference[WWW '23]{Proceedings of the ACM Web Conference 2023}{May 1--5, 2023}{Austin, TX, USA}
\acmBooktitle{Proceedings of the ACM Web Conference 2023 (WWW '23), May 1--5, 2023, Austin, TX, USA}
\acmPrice{15.00}
\acmDOI{10.1145/3543507.3583867}
\acmISBN{978-1-4503-9416-1/23/04}

\begin{document}

\title{Exploring Social Media for Early Detection of Depression in COVID-19 Patients}


\author{Jiageng Wu}
% \authornotemark{1}
\authornote{This work is done during the author’s internship at Tencent Jarvis Lab.}
\affiliation{
    \institution{Zhejiang University}
    \city{Hangzhou}
    \country{China}
}
\email{jiagengwu@zju.edu.cn}

\author{Xian Wu}
\authornote{Corresponding authors}
% \authornotemark{2}
\affiliation{
    \institution{Tencent Jarvis La  b}
    \city{Beijing}
    \country{China}
}
\email{kevinxwu@tencent.com}

\author{Yining Hua}
\affiliation{
    \institution{Harvard University}
    \city{Boston}
    \country{USA}
}
\email{yining_hua@hms.harvard.edu}

\author{Shixu Lin}
\affiliation{
    \institution{Zhejiang University}
    \city{Hangzhou}
    \country{China}
}
\email{l_sx@zju.edu.cn}

\author{Yefeng Zheng}
\affiliation{
    \institution{Tencent Jarvis Lab}
    \city{Hong Kong}
    \country{China}
}
\email{yefengzheng@tencent.com}

\author{Jie Yang}
% \authornote{2}
\authornotemark[2]
\affiliation{
    \institution{Zhejiang University}
    \city{Hangzhou}
    \country{China}
}
% School of Public Health and the Second Affiliated Hospital, 
\email{jieynlp@gmail.com}


\begin{abstract}
The COVID-19 pandemic has caused substantial damage to global health. Even though three years have passed, the world continues to struggle with the virus. Concerns are growing about the impact of COVID-19 on the mental health of infected individuals, who are more likely to experience depression, which can have long-lasting consequences for both the affected individuals and the world. Detection and intervention at an early stage can reduce the risk of depression in COVID-19 patients. 
In this paper, we investigated the relationship between COVID-19 infection and depression through social media analysis. Firstly, we managed a dataset of COVID-19 patients that contains information about their social media activity both before and after infection. Secondly, We conducted an extensive analysis of this dataset to investigate the characteristic of COVID-19 patients with a higher risk of depression. Thirdly, we proposed a deep neural network for early prediction of depression risk. This model considers daily mood swings as a psychiatric signal and incorporates textual and emotional characteristics via knowledge distillation. Experimental results demonstrate that our proposed framework outperforms baselines in detecting depression risk, with an AUROC of 0.9317 and an AUPRC of 0.8116. Our model has the potential to enable public health organizations to initiate prompt intervention with high-risk patients.

\end{abstract}

\begin{CCSXML}
<ccs2012>
    <concept>
       <concept_id>10002951.10003227.10003351</concept_id>
       <concept_desc>Information systems~Data mining</concept_desc>
       <concept_significance>500</concept_significance>
       </concept>
   <concept>
       <concept_id>10010405.10010444.10010449</concept_id>
       <concept_desc>Applied computing~Health informatics</concept_desc>
       <concept_significance>500</concept_significance>
       </concept>
 </ccs2012>
\end{CCSXML}

\ccsdesc[500]{Information systems~Data mining}
\ccsdesc[500]{Applied computing~Health informatics}


\keywords{Natural language processing, Social media, Depression detection}


% \received{18 Nov 2022}
% \received[revised]{xx xxx xxxx}     
% \received[accepted]{xx xxx xxxx}

\maketitle

\section{Introduction}
Since the outbreak of COVID-19 in 2020, this global pandemic has caused 625 million infections and 6.57 million deaths.\footnote{https://covid19.who.int/} Even though it has been three years, COVID-19 has not been eradicated worldwide. The rapid spread of the pandemic and the resulting economic downturn have exacerbated widespread anxiety, confusion, emotional isolation, and panic~\cite{mental-NEJM-2020}. According to Global Burden Disease (2020) study \cite{2020GBD-Lancet-2021}, the COVID-19 pandemic has caused nearly a 27.6\% increase in depression and a 25.6\% increase in anxiety worldwide. 

Depression, which affects an estimated 3.8\% of the world's population\footnote{https://www.who.int/news-room/fact-sheets/detail/depression}, is now the leading cause of mental health-related disease burden globally~\cite{depression-burden-lancet-2019}. Depression causes persistent feelings of sadness that negatively affect how individuals feel, think, and act. In severe cases, depression can lead to suicide \cite{suicide-depression-lancet-2016}. Approximately 5\% of depressed adolescents will commit suicide \cite{depression-suicide-intervention-1998}. However, depression is preventable and treatable \cite{depression-treatable-2010}, and the sooner it is treated, the better the outcome \cite{depression-suicide-intervention-1998}. Despite a 41\% increase in the burden of mental disorders over the past two decades \cite{suicide-depression-lancet-2016}, mental health remains one of the most neglected yet crucial development issues. In many low- and middle-income countries (LIMCs), there are fewer than one mental health worker for every 100,000 people, and more than 75\% of people do not receive treatment \cite{2020GBD-Lancet-2021}.$^2$

\begin{figure*}[ht]
  \centering
  % \includesvg[width=18cm]{Figures/early_detection_of_depression.svg}
  \includegraphics[width=18cm]{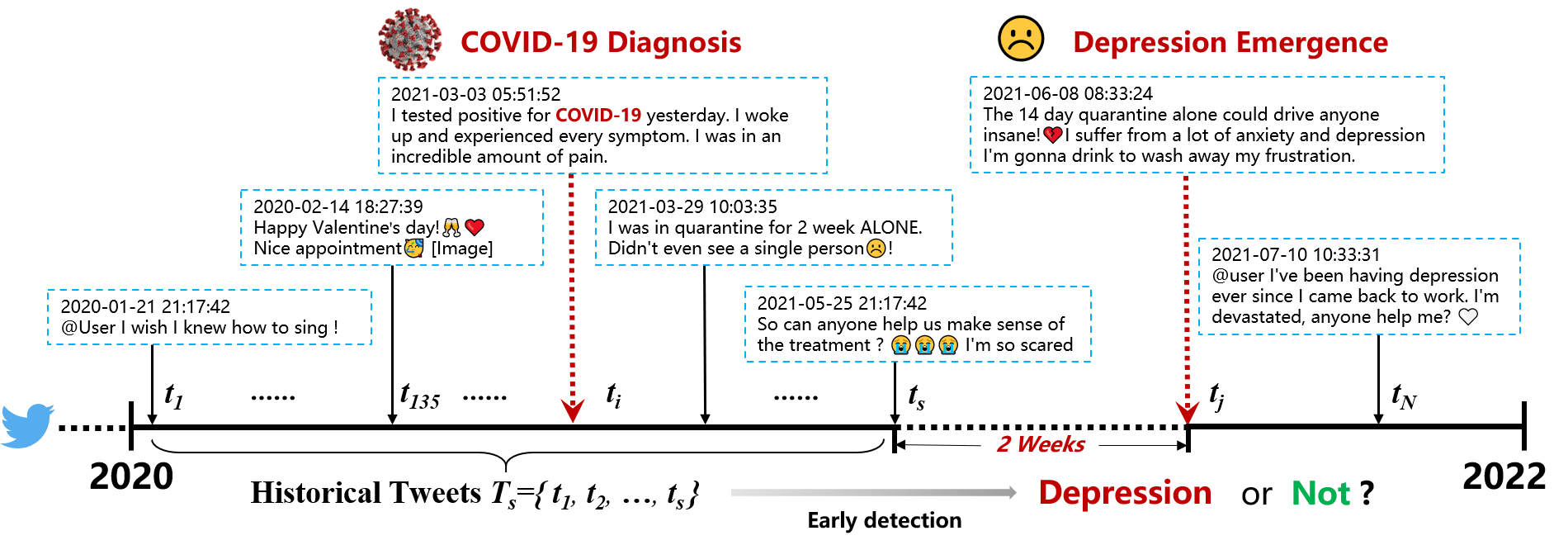}
  \caption{Tweet Timeline of a COVID-19 Patient with Depression. Tweet $t_i$ represents the patient's mention of their COVID-19 diagnosis, while tweet $t_j$ represents their mention of becoming depressed. To perform early prediction of depression risk, we selected a set of tweets $T_S = \{t_1, t_2, \ldots, t_s\}$ posted at least two weeks prior to $t_j$ to predict the likelihood of the patient developing depression. (Note that all raw tweets included in this paper have been rephrased for desensitization and brevity.)}
  \Description{early detection of depression}
  \label{fig:example}
\end{figure*}

To alleviate the depression crisis caused by COVID-19, it is crucial to detect depressed patients at an early stage so that they can receive prompt treatment \cite{benefit-early-2016}. Nonetheless, social stigma and self-stigma have emerged as significant barriers to treatment \cite{stigma-barrier-2014, stigma-barrier-2006}. Despite the fact that depression can result in social withdrawal and isolation, many affected individuals attempt to disclose their experiences on social media due to the virtuality and privacy of social identity \cite{disclose-2020, disclose-plosone}. Moreover, online communities provide a hospitable environment that enables individuals to connect with others who face comparable challenges \cite{disclose-AAAI}. After the outbreak of COVID-19, the use of social media platforms has increased by 61\% as people rely on them to stay in touch with others \cite{covid-more-social}. As more individuals with depression tend to self-disclose and seek assistance on social media, these platforms provide a rich ecosystem for studying the manifestation and characteristics of depression. 

This paper aims to develop a social media-based depression early detection model among COVID-19 patients. Using a knowledge distillation framework, our proposed model combines the longitudinal contextual information from Twitter posts and the daily emotional status of COVID-19 patients to predict their risk of depression. The contributions of this work are as follows:

Firstly, we managed a dataset (DepCOV) comprising 10,656 Twitter users. It includes users at risk for depression following a COVID-19 diagnosis and a control group. We collected the date of COVID-19 infection, pre-infection posts, and post-infection posts for each patient in the dataset. 

Secondly, we conducted in-depth experiments and data analysis to investigate the relationship between COVID-19 infection and depression. Our analysis focuses on identifying linguistic differences between depressed users and controls, as well as pre-and post-infection differences. 

Thirdly, we developed an early depression risk detection model for COVID-19 patients. Figure \ref{fig:example} illustrates the historical tweets of a COVID-19 patient. The patient was infected around the time of tweet $t_i$ and developed depression signals around tweet $t_j$. To perform early prediction of depression risk, we selected tweets posted at least two weeks prior to $t_j$. Given the significant negative impact of COVID-19 on mood \cite{covid-mood-swing-2021, covid-mood-swing-2022}, we used mood swings as a potential diagnostic signal for depression detection. Our proposed Mood2Content model integrates both textual and emotional features through knowledge distillation to make predictions. Experiment results show that Mood2Content outperforms other competitive baselines, achieving high performance with an AUROC of 0.9317 and AUPRC of 0.8116. 

\section{RELATED WORK }
\subsection{Depression Detection in Social Media}
Unlike the conventional machine learning task in other fields that are supported by extensive and high-quality datasets with gold-standard diagnoses, the myriad of privacy and ethical concerns of mental disorders have limited the accessibility of datasets with clinically validated diagnostic information. Consequently, many researchers have devoted themselves to constructing reliable datasets to support various tasks. The annotation/development schemes of a dataset are mainly based on affiliation behaviors, self-reports, and expert/external validation (see more details in \cite{review-nDM-2020, dataset-gap-CHI}). The most ideal datasets are curated by the third scheme, which introduces the experts' examination \cite{dataset-psychiatrist-2017} or incorporate electronic healthcare records \cite{dataset-ehr-facebook-2018}, but its effort- and time-consuming nature limit its scale, diversity, and accessibility \cite{dataset-gap-CHI}. Therefore, the first two methods are the most popular and practical schemes. The first strategy operationalizes hashtags, account following, and community participation related to psychiatric resources as interested signals, such as followers of psychiatrist account \cite{dataset-follower-2015}, posts in depression forum \cite{dataset-forum-2014, dataset-forum-2017, dataset-TRT-2018}. The third scheme identifies the interested person according to their self-disclosure in social media, such as the matching pattern for feelings or diagnoses of mental disorders (e.g., \textit{"I was diagnosed with depression"}). For example, \cite{dataset-self-1-2014} adopted the regular expression of diagnosed pattern to seek persons with mental disorders in Tweet. Since then, more similar datasets have been proposed, such as RSDD \cite{dataset-RSDD-2017}, SMHD \cite{dataset-SMHD-2018}, eRisk \cite{dataset-erisk-2019}, and have flourished related workshops, such as CLPsych \cite{competition-clpsych-2015} and eRisk \cite{competition-erisk-2017}. Beyond these, Kelly and Gillan recruited participants who self-reported depressive episodes through an online worker platform \cite{dataset-recruitment-NC-2022, dataset-recruitment-nDM-2022}. 

Research about mental health based on social media mainly focused on the detection model and the potential indications of mental disorders. For model development, the classical paradigm is the combination of feature extraction and classifier, such as linguist features with logistic regression \cite{dataset-self-1-2014}. The common feature extraction methods include TF-IDF, word embedding \cite{depression-word-embedding-2018}, LIWC (Linguistic Inquiry and Word Count) \cite{method-liwc-2010}, and LDA (Latent Dirichlet Allocation) \cite{lad-2003}. Currently, more research has gradually used deep learning models to represent posts, including convolution neural network (CNN) \cite{dataset-SMHD-2018}, recurrent neural network (RNN) \cite{method-RNN-2018}, long short-term memory neural network (LSTM) \cite{method-LSTM-2019} and Transformer \cite{method-transformer-abstract-2021}. Meanwhile, \cite{method-multimodality-gui-2019} and \cite{method-multimodality-An-2020} cooperated with the text and image by the multi-modality model. Especially, several research introduced the attention-based approach to improve model interpretability and generalizability \cite{method-SS3-2019}, such as hierarchical attention networks (HAN) \cite{method-HAN-2016}. And, recent studies cooperated with the psychiatric scale of clinical diagnosis to guide depression detection \cite{method-scale-acl-2022, method-scale-ijcai-2022}. 

There is a growing interest in exploring the potential of social media for depression diagnosis, including linguistic characteristics \cite{method-semantic-2022, dataset-ehr-facebook-2018} and social behavior \cite{method-behavior-2017}. Studies have shown that LIWC, LDA, and text clustering can be used to examine linguistic differences between individuals with schizophrenia and healthy controls \cite{method-linguistic-schizophrenia-2015}. Trotzek et al. \cite{depression-word-embedding-2018} built a logistic regression classifier by integrating readability and emotion features into user-level linguistic metadata and further improved it with a CNN-based model. The work in \cite{method-multimodality-shen-2017} involved depression detection using multi-modality features such as social network features, user profile features, visual features, emotion features, topic-level features, and domain-specific features. Yang et al. \cite{mental-knowledge-2022} extracted mental state knowledge and infused it into a GRU model to explicitly model the mental states of the speaker. Kelley et al. \cite{dataset-recruitment-NC-2022} constructed personalized, within-subject networks based on depression-related linguistic features from LIWC and discovered a positive correlation between overall network connectivity and depression severity. The negative mood, a typical symptom of depression, has also been extensively studied in the context of social media posts, with the majority of research concentrating on content analysis or the extraction of hand-crafted features using lexicons or rules \cite{depression-sentiment-lexicon-2013, depression-sentiment-lexicon-2021}.

% \subsection{Learning Models for COVID-19}
As a global health crisis, COVID-19 has received significant attention on social media. On the basis of large-scale social media data, there has been an abundance of research on COVID-19 \cite{social-covid-lancet-2021}, including thematic analysis \cite{covid-lda-2020}, symptom identification \cite{covid-symptom-2020, symptom-social-covid-2022}, and public perception analysis \cite{covid-perception-2020, mh-covid-2022}. However, research modeling the relationship between depression and COVID-19 is scarce. This paper represents, to the best of our knowledge, the first attempt to predict early the depression risk of COVID-19 patients.

\subsection{Research on Knowledge Distillation}
Large-scale deep learning models have limited practical applications due to their computational complexity and storage requirements. Knowledge distillation (KD) is a solution to this issue, as it enables the distillation of a large model into a smaller model with a relatively low reduction in performance \cite{distll-review-2021}. The
student model of KD is synchronously guided by the distillation loss that reflects the gap between the student model and teacher model, and the task loss that measures the prediction errors of the student model \cite{distill-hinton-2015}. Different KD strategies define distillation loss differently. Distilled BiLSTM \cite{distill-distilled-BiLSTM-2019} used the MSE loss between the output of the teacher model and the student model as the distillation loss. BERT-PKD \cite{distll-pkd-2019} extracted information from intermediate layers and computed the MSE loss. DistillBERT \cite{distill-distillBERT-2019} and TinyBERT \cite{distill-tinyBERT-2019} guided the student model in the pre-training stage. MiniLM \cite{distill-minilm-2020} further distilled the self-attention distributions and value relations of the teacher's last Transformer layer to guide the student model training, making it effective and generative for student models.

\section{METHODOLOGY}

\begin{figure*}[ht]
  \centering
  \includegraphics[width=12cm]{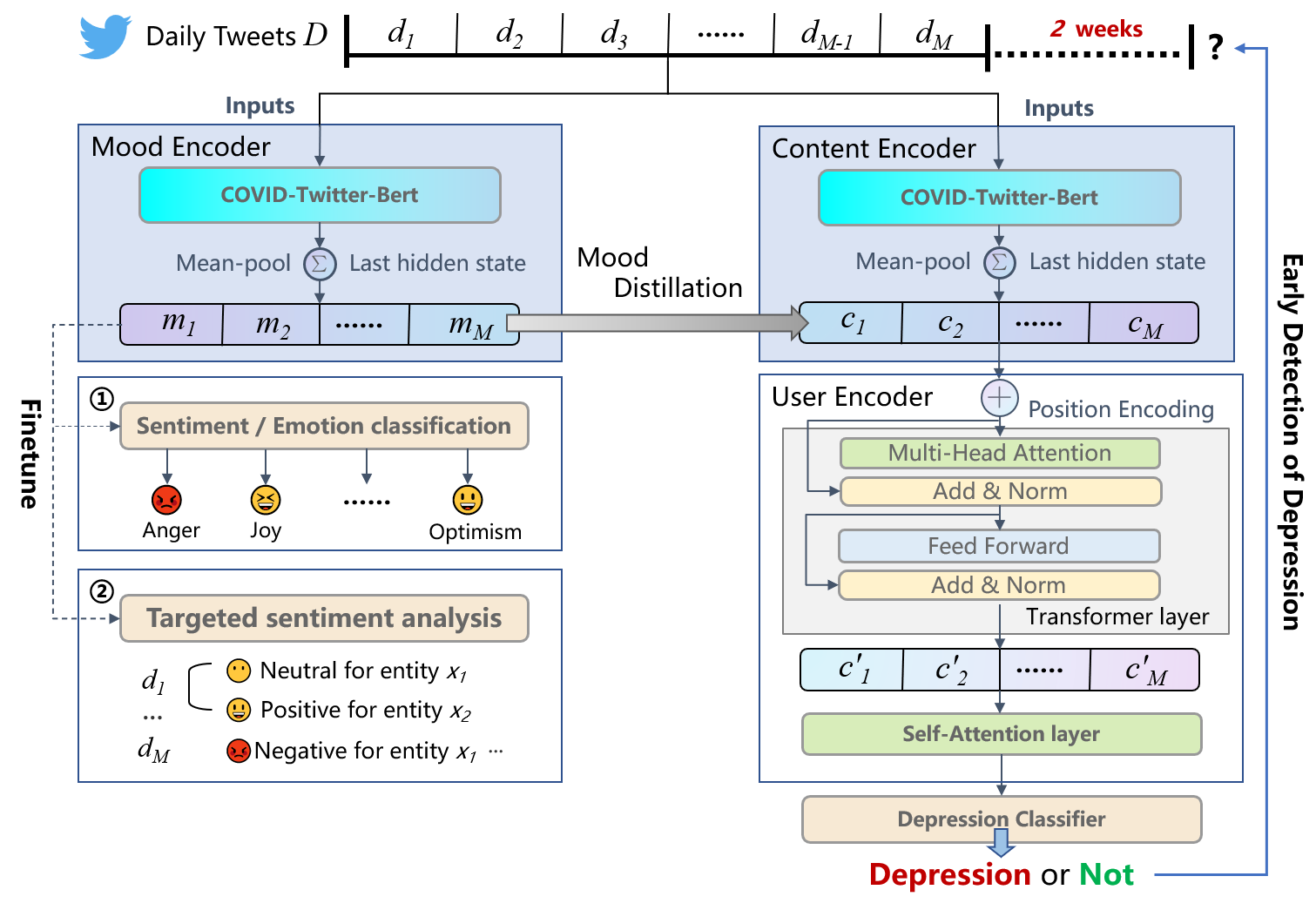}
  \caption{Framework of the Mood2Content model which includes a Content Encoder and a Mood Encoder, both encoders accept daily aggregated tweets as input and target to predict the depression risk at an early stage.}
%   \Description{Mood2Content model}
  \label{fig:framework}
\end{figure*}

\subsection{Problem Formulation}
This study uses Twitter as the major social platform to detect depression and predict early-stage risks. Given a user $u$, we can acquire his historical tweets, such as posts and comments, which contain abundant information about personal experiences and feelings. We denote all tweets acquired from $u$ with $T=\{t_1, t_2, \ldots, t_N\}$, where $N$ is the total number of tweets. We also denote the first tweet mentioning being infected by COVID-19 as $t_i$ and the first tweet emitting depression signals as $t_j$. In this paper, we aim to detect potential depression before users explicitly express depressive feelings and after they get COVID-19. Therefore, we focus on cases where $t_j$ is posted after $t_i$. For early detection, we further limit our study range to tweets posted at least two weeks before $t_j$. Consequently, the early depression risk prediction problem can be formulated as a binary classification problem on predicting a future depression label $y$ for the user $u$ using the subset $T_s=\{t_m, t_{m+1}, \ldots, t_n\}$ from $T$, where $t_i < t_m < t_n < t_j$ on timescale.

\subsection{Feature Extraction}

\subsubsection{COVID-19 Infection Time Extraction}
This subsection presents the extraction of $t_i$, i.e., the first tweet where the user self-reported a COVID-19 diagnosis. We identify self-reported COVID-19 tweets through the following steps: 1) use keywords to filter tweets that contain specific expression phrases, such as “get COVID”, or “test positive”. Then, we use dependency parsing (supported by Stanza \cite{tools-staza-2020}) and rule-based approaches (such as negation detection) to determine the subject of infection. The first tweet with the user as the infection subject is associated with a timestamp, but this timestamp does not necessarily represent the user's infection time $t_i$. Therefore, we further applied the regular expression to extract time information in the tweet to determine the user's infection time $t_i$. More details and related resources on dataset construction can be found in the code repository.\footnote{https://github.com/Dragon-Wu/DepCov-WWW2023} 

\subsubsection{Depression Time Extraction}
This subsection presents the extraction of $t_j$, i.e., the first tweet where the user expressed depressed feelings after COVID-19 infection. Following \cite{dataset-SMHD-2018, method-scale-acl-2022}, we define self-reported depression as tweets that mention depression conditions and first-person pronouns within a short lexical distance. Based on official psychiatric resources\footnote{https://www.mayoclinic.org/diseases-conditions/depression/}$^,$\footnote{https://www.who.int/news-room/fact-sheets/detail/depression}$^,$\footnote{https://www.nimh.nih.gov/health/topics/depression}, we curate a comprehensive lexicon of depression conditions. The lexicon contains various expressions of depressive disorders (e.g., \textit{major depression disorder, dysthymia}), the status of extreme depression mood (e.g., \textit{miserable, hopeless}), and typical symptoms of depression (e.g., \textit{suicide, severe mood swings}). In addition, we also add colloquial expressions. With this lexicon and high-precision regular expression, we extract tweets with depressive signals and remove tweets with ambiguity, non-self-report, and negation. Manual validation on a random sample of 200 tweets shows an accuracy of 91.0\%. 

\subsubsection{Aggregation of Daily Tweet} 
After identifying the infection time $t_i$ and depression tweet $t_j$ were identified, we selected all tweets that were posted before $t_j$ (e.g., two weeks) and denoted them as $T_s={t_m, t_{m+1}, \ldots, t_n}$. The objective was to extract features from $T s$ in order to predict whether this user would develop depression in the near future. Due to Twitter's character limit, tweets were typically brief, making semantic and sentiment analysis difficult and resulting in frequent mood swings. To address this issue, we condensed the historical tweets $T_s$ into daily tweets $D={d_1, d_2, ..., d_M}$ and sorted them in reverse order, where $d_i$ represented all tweets generated on the $i$th day. While everyone's mental state fluctuates over time, including those of depressed patients, using tweets posted a long time ago may not have accurately reflected their current mental state and could have led to inaccurate predictions. To improve efficiency and focus on the current state of the user, we truncated historical posts after four weeks, enabling online and timely detection. Consequently, the maximum number of elements in $D$ was at most 28 (4 $\times$ 7). If there are not 28 daily tweets from the past four weeks, the latest daily tweets from historical tweets will be collected to meet the sliding window.

\subsubsection{Tweet Representation}
After merging daily tweets into $D=\{d_1, d_2, ..., d_M\}$, we adopt BERT \cite{method-BERT-2018} as the textual encoder to represent each $d_j$ in $D$. Here we use the COVID-Twitter-BERT-v2 (CTB) \cite{BERT-covid-2020}, a BERT-large-uncased model that has been incrementally pre-trained on large-scale COVID-19-related tweets. In recognition of the important role of emotional information in depression detection, we also develop a Mood Encoder to capture the emotional context of tweets. To enhance its capability, we further pre-train the CTB model on three sentiment-related tasks: sentiment classification \cite{encoder-tweeteval-2020, semeval-sentiment-2017}, emotion recognition \cite{semeval-emotion-2018}, and targeted sentiment analysis \cite{encoder-metscov-2022}. These three tasks yield three optimized models based on CTB, we denote them with CTB-St, CTB-Emo, and CTB-Tsa, respectively.

The CTB-St and CTB-Emo models were fine-tuned using the SemEval 2017-Sentiment Analysis in Twitter \cite{semeval-sentiment-2017} and SemEval 2018 - Emotion Recognition \cite{semeval-emotion-2018} datasets, respectively. CTB-St categorizes the overall sentiment of tweets into negative, neutral, and positive, while CTB-Emo infers the emotional state of a tweet (anger, joy, sadness, optimism). Both models were fine-tuned with a basic BERT setting, which involves mean-pooling the embeddings of the last hidden state of CTB and inputting it into a linear classifier. The third task, TSA (Targeted Sentiment Analysis), is a fine-grained sentiment analysis aimed at inferring user sentiment toward targeted entities (negative, neutral, and positive). The CTB-Tsa model was fine-tuned on the METS-CoV dataset \cite{encoder-metscov-2022}, which contains COVID-19 related tweets, using the BERT-SPC model setting \cite{method-BERT-2018}.

For each daily aggregated tweet $d_j=\{w_{1}, w_{2}, ..., w_{N_j}\}$ of user $u$, we adopt the mean-pooling of the last hidden state of BERT model as the tweet representation: 
\begin{equation}
  c_j = Content_{Encoder}(d_j) =\frac{1}{N_j}\sum^{N_j}_{l=1}BERT_{|LAST|}(w_{1}, w_{2}, ..., w_{l}) 
\end{equation}
where $c_j$ refers to content representation of $d_j$.
\begin{align}
  m_j & = Mood_{Encoder}(d_j) 
  \label{eqn:mood}
\end{align}
where $m_j$ refers to the mood representation of $d_j$ and $Mood_{Encoder}$ can be one of CTB-St, CTB-Emo and CTB-Tsa.

\subsection{Mood2Content Model}
As shown in Figure \ref{fig:framework}, we propose a novel framework Mood2Content that cooperates with both the content representation and mood representation to conduct early detection of depression.
Given a user $u$ to with daily merged tweets $D=\{d_1, d_2, ..., d_M\}$, we can use the content encoder and the mood encoder to acquire the corresponding representation $C= \{c_1, c_2, \ldots, c_M\}$ and $M= \{m_1, m_2, \ldots, m_M\}$. Then we generate the embedding $x$ of user $u$ based on $C$ and $M$. 

The set $D$ contains the merged daily posts sorted in reverse chronological order, such ranking information needed to be included in modeling. This is because the most recent tweets record the current status of this user which is more informative for future depression risk prediction. As a result, we add the position information to the content representation $c_j$ of each $d_j$ in $D$. 
\begin{equation}
  c'_{j} = Concatenation(c_j,pos_j)
\end{equation}
where $pos_j$ is a hard position embedding denoting the day gap between the $j$th day and now, emphasizing the timeline information. 

After updating content representation with position information, we acquire the user representation $x$ with a user encoder that consists of Transformer and self-attention layer. Transformer enables $c'_j$ to utilize the information from other daily tweets. $s_j$ is the $j$th embedding of the last hidden state of Transformer:
\begin{equation}
  s_{j} = User_{Encoder|LAST|}(C', j)
\end{equation}
A self-attention layer is used to generate the weighted sum of all $s_i$:
\begin{equation}
  \alpha_j = \frac{exp(Ws_{j}+b)}{\sum^M_{k=1}exp(Ws_{k}+b)}
\end{equation}
\begin{equation}
  x = \sum_{j=0}^{M}\alpha_js_{j}
\end{equation}
where $W$ are learnable parameters.
Then, the user representation $x$ is the input of the classifier head $F$ (linear layer) to predict the depression risk $p$.
\begin{equation}
  p = SIGMOID(W_F\cdot x + b_F)
\end{equation}
where $W_F$ are learnable parameters.
Therefore, the model can be trained with the loss function of depression prediction $\mathcal{L}_{clf}$ :
\begin{equation}
  \mathcal{L}_{clf}=y\cdot log p + (1-y)\cdot log(1-p)
\end{equation}

% Notably, if we directly adopt the emotional BERT model (CTB-St, CTB-Emo, CTB-Tsa) as the only post encoder, which seem to model the changes of mood contained in daily posts as the diagnostic signal for depression. 

% \textbf{Mood2Content Model:} 
To integrate the mood representation into depression risk prediction, inspired by knowledge distillation, we guide the content encoder to align with the mood representation. In detail, we first acquire the mood representation $M= \{m_1, m_2, \ldots, m_M\}$ in Eq.(\ref{eqn:mood}). Then, this mood encoder will be frozen in depression detection and no longer update model weights. We introduce $\mathcal{L}_{distill}$ as a distance measure between mood vector $M_i$ and content vector $C_i$, which guides the content encoder to reach a trade-off between feature fusion and model classification: 
\begin{equation}
  \mathcal{L}_{distill}=\lvert\lvert M - C \lvert\lvert^2_2
\end{equation}
Therefore, the Mood2Content model is optimized towards both mood distillation and prediction error reduction. The overall loss of model can be formulated as a weighted sum of $\mathcal{L}_{clf}$ and $\mathcal{L}_{distill}$:
\begin{equation}
\begin{split}
  \mathcal{L}&=\alpha \cdot \mathcal{L}_{clf} + (1-\alpha) \cdot \mathcal{L}_{distill} \\
  &=\alpha \cdot [y\cdot log p + (1-y)\cdot log(1-p)] + (1-\alpha) \cdot \lvert\lvert M - C \lvert\lvert^2_2
\end{split}
\end{equation}
where $\alpha$ is an adjustable factor that can emphasize feature fusion or classification. 

\section{EXPERIMENTS}
\subsection{Dataset}
We select original English tweets related to COVID-19 using unique tweet identifiers (tweet ID) from a widely used open-source COVID-19 tweet database \cite{model-datacollection-chen-2017, model-datacollection-lopez-2021}. These tweets were identified by Twitter’s trending topics and keywords associated with COVID-19, such as \textit{COVID-19} and \textit{SARS-COV-2}. We first download 471,553,966 target tweets across 27 months, from February 1st, 2020, to April 30th, 2022, using Twitter’s Application Programming Interface (API). After the identification of COVID-19 patients, we further collect retrospective tweets between January 1st, 2020, and December 31st, 2021 from each infected user for further analysis and modeling. 

Due to the mental disease problems brought by COVID-19, we presume that there are many vulnerable persons who may present depression risk after COVID-19 diagnosis. We split the entire user set into two groups according to their quantity and the corresponding timestamp of depression tweets: 1) the first group is the treatment group which includes users emitting depression signals after suffering COVID-19. We require users in this group to have posted more than three depression tweets and the first of which was posted at least two weeks after their COVID-19 diagnosis. In addition, these users never post a depression tweet before COVID-19. Particularly, we set a window period of two-week, a widely used time window in the diagnosis of mental disorders, between COVID-19 diagnosis and the emergence of depression risk and the subsequent modeling and analysis merely utilize their tweets before it; 2) the second group is the control group which includes users who don't mention depression both before and after COVID-19 infection. For each user in the first group, we select 5 users with a similar quantity of tweets and add them to the second group. Besides, all eligible users must contain more than 25 tweets both before and after the COVID-19 diagnosis respectively, $\geq75\%$ of which are written in English. 

In this manner, we build a dataset of COVID-19 patients with depression signals and name it the DepCOV dataset. DepCOV consists of 1,776 depression cases (positive) and 8,880 controls (negative), with 10,488,061 tweets. For model development and evaluation, We split the DepCOV into the training set, validation set, and testing set with the proportion of 7:1:2. 

As the overall statistic of DepCOV is shown in Table \ref{Table:DepCOV}, the depressed person among COVID-19 patients posted more tweets than the controls, and this has been further enhanced after they got COVID-19. Besides, the depressed users in the DepCOV have an average of 5.27 depression tweets and the time between their COVID-19 diagnosis and depression was an average of 59.41 days. 

\subsection{Settings}

\begin{table}[t]
\caption{The statistics of the proposed dataset DepCOV. (Both include COVID-19 patients who will or will not get depressed in two weeks)}
\begin{center}{
\resizebox{\linewidth}{!}{
\begin{tabular}{ c  c  c  c  c  c }
\hline
\multicolumn{1}{c}{{\textbf{Statistics}}} & \multicolumn{2}{c}{\textbf{Depression (n=1,776)}} & \multicolumn{2}{c}{\textbf{Controls (n=8,880)}} & \multicolumn{1}{c}{\textbf{DepCOV (n=10,656)}} \bigstrut\\
\cline{2-6} \multicolumn{1}{c}{\textbf{(Mean)}} & \multicolumn{1}{c}{\textbf{Before COV}} & \multicolumn{1}{c}{\textbf{COV to Dep}} & \multicolumn{1}{c}{\textbf{Before COV}} & \multicolumn{1}{c}{\textbf{After COV}} & \multicolumn{1}{c}{\textbf{Overall}} \bigstrut\\
\hline
Tweets Count  &583.74 	&383.44 	&586.87 	&400.78 	&492.12 \\
\hline  
Days Count   &109.57 	&59.01 	&149.40 	&108.70 	&121.59 \\
\hline  
Tweet length	&23.21 	&23.95 	&21.44 	&21.47 	&21.81 \\
\hline 
Tweet per day	&5.33 	&6.50 	&3.93 	&3.69 	&4.16 \\
\hline
Daily Tweet length    &113.64 	&135.88 	&78.64	&75.86	&85.17  \\
\hline
\end{tabular}}}
\label{Table:DepCOV}
\end{center}
\end{table}

% \textbf{Hyper-parameter setting:} 
To evaluate the model performance objectively, models of the same type have exactly the same parameters. The max number of training epochs is 10 and the patience of early stop is 10. The training batch size is 32 and the learning rate is 5e-5 with the cosine scheduler with warm-up. 
% The user encoder of HAN-User and Mood2Content have 12 Transformer layer with 16 attention heads. 
The $\alpha$ of Mood2Content model is 0.5, which yields a balance between distillation loss and classification loss. 

% \textbf{Metrics:} 
To avoid the influence of randomness, we run each model with 3 different seeds (42, 52, 62) and report the average performance. For the practical availability and generalizability, we adopt the area under the receiver operating characteristic curve (AUROC) and the precision-recall curve (AUPRC) instead of accuracy or F1-score, which are widely used in such tasks but set a hard threshold of 0.5. AUROC and AUPRC can more comprehensively evaluate the model performance regardless of any threshold, enabling more aggressive or conservative interventions for persons at depression risk. 

\subsection{Analysis}
\subsubsection{Linguistic discrepancy}
To analyze content differences, we compared psycholinguistic characteristics between COVID-19-infected patients who developed depression and those who did not, as well as between tweets posted by depressed patients prior to and after their COVID-19 diagnosis. We utilized the LIWC lexicon, a psychometrically validated mapping of words to psychological concepts that had been widely applied to the analysis of mental health in social media text  \cite{method-liwc-2010, dataset-self-1-2014}. We conducted Chi-square tests on each characteristic and determined its odds ratios (ORs). Table \ref{Table:LIWC} displayed the characteristics with the ten highest and ten lowest odds ratios, with p-values for each result <0.0001.

Compared to pre-COVID-19 diagnosis, depressed individuals used fewer words associated with recreation (leisure, positive emotion, friends, and motion) and more words associated with sexuality, health, risk, negation, and anger, indicating a change in their lifestyle and concerns \cite{depression-creation-2021}. Similarly, depressed individuals expressed fewer positive words (accomplishment, reward, power, and leisure) than non-depressed individuals. Specifically, and in accordance with clinical or social media studies, we observed that depressed individuals tended to use more first-person pronouns than controls, indicating an increase in self-focused attention \cite{dataset-recruitment-nDM-2022, depression-firstperson-2017}. In addition, an increase in female- and family-related words may reflect a description of familial affection.

\begin{table}[t]
    \caption{Discrepancy of psycholinguistic feature. \\(The odds ratios (ORs) quantify the linguistic disparities between depression and controls, as well as between the pre- and post-COVID phases of the depression. All \textit{p} < 0.0001)}
    \centering
    \resizebox{\linewidth}{!}{
    \begin{tabular}{c c c c c c c c}
    \hline
        \multicolumn{4}{c}{\textbf{After COVID-19 VS Before COVID-19 (Depression)}} & \multicolumn{4}{c}{\textbf{Depression VS Controls}} \\ \cmidrule(r){1-4} \cmidrule(r){5-8}
        \textbf{Category} & \textbf{OR} & \textbf{Category} & \textbf{OR} & \textbf{Category} & \textbf{OR} & \textbf{Category} & \textbf{OR} \\ \hline
        Leisure & 0.92 & Sexual & 1.08 & Money & 0.85 & I & 1.26 \\ \hline
        Ingest & 0.94 & Health & 1.04 & You & 0.88 & Female & 1.17 \\ \hline
        Nonfluencies & 0.96 & Risk & 1.04 & Achievement & 0.93 & Family & 1.10 \\ \hline
        See & 0.96 & They & 1.04 & Work & 0.93 & Ingest & 1.10 \\ \hline
        Home & 0.97 & Money & 1.04 & Death & 0.93 & Filler Words & 1.10 \\ \hline
        Affiliation & 0.97 & Causal & 1.03 & Reward & 0.93 & Anxiety & 1.08 \\ \hline
        Positive Emotions & 0.97 & SheHe & 1.03 & Power & 0.94 & Insight & 1.08 \\ \hline
        Perceptual Processes & 0.97 & Negations & 1.02 & Leisure & 0.94 & Feel & 1.07 \\ \hline
        Friends & 0.97 & Insight & 1.02 & We & 0.96 & Assent & 1.06 \\ \hline
        Motion & 0.97 & Anger & 1.02 & Drives & 0.96 & Religion & 1.05 \\ \hline
    \end{tabular}
    \label{Table:LIWC}
    }
\end{table}

\subsubsection{Content analysis of Depression tweet} 
To shed light on the potential underlying causes of depression, we analyzed the content of tweets pertaining to depression. The tweets were filtered for depression-related conditions, and Latent Dirichlet Allocation (LDA) was used to identify the primary concerns of depressed individuals and determine what factors may have contributed to their depressive moods. The number of topics was limited to between ten and 200, and the optimal model was chosen based on its coherence and complexity. Table \ref{Table:LDA} displays the leading ten topics and their top 20 words for the best model, which had 125 topics. Our analysis revealed that the sources or targets of negative emotions were frequently associated with the ongoing pandemic, such as the lockdown, government, treatments, and the disease itself. In addition, additional topics centered on the participants' emotions and feelings, including depression, encouragement, and complaints.

\begin{table}
  \caption{The most concerned topics of depression tweets (Top-10 words of top-10 topics)}
  \resizebox{\linewidth}{!}{
  \begin{tabular}{cc}
    \toprule
    \textbf{Topic} & \textbf{Keywords}\\
    \midrule
    Lockdown & \makecell[c]{absolutely, outside, door, figure, stream, \\breathe, pressure, air, strange, day} \\
    \hline
    Government & \makecell[c]{place, government, wonder, explain, bunch, \\result, people, fix, citizen, believing} \\
    \hline
    Depression & \makecell[c]{attack, panic, panic\_attack, piece, awful, \\reminds, time, victim, forced, failure} \\
    \hline
    Policy & \makecell[c]{American, fast, freedom, middle, exist, \\accept, overwhelming, hero, military, ocd} \\
    \hline
    Encouragement & \makecell[c]{went, fall, strong, time, happens, \\option, praying, counseling, stay, stay\_strong} \\
    \hline
    Complaint & \makecell[c]{mind, fear, fact, past, win, \\space, city, committing, medicine, the\_fact} \\
    \hline
    Treatment & \makecell[c]{second, heard, happened, sadness, pill, \\smoking, xanax, intense, recovery, nausea} \\
    \hline
    Disease & \makecell[c]{heart, ask, important, ill, beat, \\present, reality, pm, heart\_attack, alcohol} \\
  \bottomrule
\end{tabular}
\label{Table:LDA}
}
\end{table}

\subsection{Early Depression Detection }

\subsubsection{Baselines} 
To fully evaluate the performance of different models in our experiment settings, we constructed several baselines which range from statistical NLP models to deep learning models. 

\textbf{LIWC+LR:} As users' language characteristic can reveal their psychological state, the first baseline is LIWC+LR which extracts the psycholinguistic features of merged historical tweets by LIWC \cite{method-liwc-2010} and predict the depression risk by logistic regression classifier. 

\textbf{TF-IDF+XGBoost:} It adopts TF-IDF weighted features of word and character n-grams and the popular machine learning model XGBoost \cite{model-xgbboost-2015}, which is also extensively used in similar tasks \cite{dataset-TRT-2018}. 

\textbf{HAN}: The simple concatenation of all tweets in the absence of temporal information may easily lead to the crucial clues lost in large-scale corpus \cite{method-scale-ijcai-2022}. Therefore, we select the representative HAN \cite{method-HAN-2016} as a deep learning baseline, which conducts depression prediction with a hierarchical attention neural network. HAN obtains each tweet $t_j$ representation with bidirectional GRU and further encodes all tweet representations into user presentation with an attention mechanism to conduct the final prediction. To improve model performance, we select the 500, 1000, 2000 latest tweets, and 4-weeks' daily tweets as model input to develop baseline model \textbf{HAN-500}, \textbf{HAN-1000}, \textbf{HAN-2000} and \textbf{HAN-Daily}, respectively. \textbf{HAN-Daily(BERT)} takes daily tweets as input and replaces the BiGRU with CTB. On the basis of HAN(BERT), \textbf{HAN-User} adds the user encoder of Mood2Content to encode tweet representation. 

\textbf{Mood\&Content}: It directly concatenates the mood representation and the content representation of daily tweets to generate the combined representation, which acts as the tweet representation to generate user representation for subsequent prediction. 

\subsubsection{Results}

As the model performance shown in Table \ref{Table: performance of baseline} and Table \ref{Table: performance of ablation study}, the proposed Mood2Content outperforms other models in early depression detection. Among the baseline models, the HAN-Daily achieved the highest performance, indicating that the recent and aggregated daily tweet can promote the model to seize the recent changes in users' mental status. Meanwhile, the inclusion of more historical tweets did not always improve performance by comparing the results of HAN-500, HAN-1000, and HAN-2000.

For the advanced model, extensive experiments were conducted to demonstrate the performance of the different strategies and to investigate the effects of different components simultaneously: 

\textbf{Effect of Tweet Encoder} Owing to the selection and aggregation of daily tweets, a large-scale pre-trained language model can be used as a tweet encoder instead of conventionally shallow CNN or GRU \cite{dataset-RSDD-2017}. The HAN(BERT) model achieved an improvement of 0.1333 in AUPRC and 0.0493 in AUROC at most than the original HAN with BiGRU. Meanwhile, other models with BERT-based tweet encoders all achieved good performance. 

\textbf{Effect of User Encoder} Compared with HAN(BERT), the HAN-User was improved by introducing a user encoder, which consists of position embedding, Transformer, and a self-attention layer to further encode tweet representation. The cooperation of position embedding and Transformer enables the user encoder can capture the longitudinal information and the final self-attention layer improves the model interpretability. 

\begin{table}
  \caption{Performance of fine-tuned CTB on sentiment tasks. \\(CTB-st is fine-tuned on  sentiment classification task, CTB-Emo is fine-tuned on emotion recognition task and CTB-Tsa is fine-tuned on targeted sentiment analysis task)}
  \label{tab:freq}
  \begin{threeparttable}
  \begin{tabular}{cccc}
    \hline
    \textbf{Model} & \textbf{Acc} & \textbf{Recall} & \textbf{F1} \\
    \hline
    CTB-St &0.7183 	&\textbf{0.7260} 	&0.7173  \\
    \hline
    CTB-Emo &0.8294 	&0.8034 	&\textbf{0.7974} \\
    \hline
    CTB-Tsa &76.29 	&0.6738 	&\textbf{0.7003}\\
    \hline
\end{tabular}
\begin{tablenotes}
\item \footnotesize (The bold means the recommended metric)
\end{tablenotes}
\end{threeparttable}
\label{Table: performance of finetune}
\end{table}

\textbf{Effect of Emotional Signal} With 
The result of HAN(BERT) and HAN-User demonstrated that emotional BERT can be also used as a tweet encoder, and CTB-Tsa resulted in better performance than general BERT in them. Notably, such strategy acts as modeling daily mood swing as the potential diagnostic signal for depression detection, which is consistent with psychiatric studies \cite{depression-mood-swing-2000, depression-mood-swing-2010} and the clinical practice \cite{depression-scale-phq9-2001, depression-scale-beck-1987}. 

\textbf{Effect of Different Mood Encoder} The results of fine-tuned models are shown in Table \ref{Table: performance of finetune}, which all nearly reached the reported SOTA performance. We examined the improvement brought by the different mood encoders. All the mood coders performed about the same, and the CTB-Tsa achieved the best result in 3/4 models. This could be due to the specific COVID-Twitter dataset, or the fine-grained sentiment analysis could capture more mood information. 

\textbf{Effect of Mood Distillation} As shown in Table \ref{Table: performance of ablation study}, the proposed Mood2Content yielded the highest performance with an AUPRC of 0.8116 and an AUROC of 0.9317. Compared to HAN-User, which relied solely on the content or emotional information, Mood2Content improved by an average of 0.0548 in AUPRC and 0.0160 in AUROC. However, Mood\&Content also contained the content and emotional information, it performed even worse than the model with single-resource information. This may be due to the huge gap between their semantic space because they are designed to capture the different contextual presentations. To address this discrepancy, Mood2Content guided the semantic space of the content encoder close to that of the mood encoder through knowledge distillation and learned to conduct depression detection simultaneously. 

\begin{table}
  \caption{Performance of early detection of depression with different models and strategies.\\( Mood\&Content is a simple concatenation of mood and content representation. Mood2Content is the proposed model which guides content representation with mood representation.)}
  \begin{threeparttable}
  \begin{tabular}{ccc}
    \hline
    \textbf{Model} & \textbf{AUPRC} & \textbf{AUROC} \\
    \hline
    LIWC+LR & 0.2815 & 0.7017 \\
    \hline
    TF-IDF+XGBoost &0.4737 &0.7933 \\
    \hline
    HAN-500 tweets &0.3219  &0.7082  \\
    HAN-1000 tweets &0.3269  &0.7138  \\
    HAN-2000 tweets &0.2623   &0.6251  \\
    HAN-Daily &0.6026 &0.8621 \\
    HAN-Daily(BERT)* &0.7359   &0.9114 \\
    \hline
    HAN-User* &0.7649   &0.9198 \\
    \hline
    Mood\&Content* &0.5364   &0.8447 \\
    \hline
    Mood2Content* &\textbf{0.8116}   &\textbf{0.9317} \\
    \hline
\end{tabular}
\begin{tablenotes} 
\footnotesize 
\item *means the best performance was reported
\end{tablenotes}
\end{threeparttable}
\label{Table: performance of baseline}
\end{table}

\begin{table}[t]
\caption{Ablation study. \\(HAN-Daily is the basic content encoder w/o user encoder and mood representation; HAN-User includes the user encoder; Mood\&Content concatenates mood and content representation. Mood2Content is the proposed model.)}
\resizebox{\linewidth}{!}{
\begin{tabular}{ccc}
    \hline
        \textbf{Model - Encoder} & \textbf{AUPRC} & \textbf{AUROC}  \\
        \hline
        HAN-Daily(BERT) - CTB & 0.6794 & 0.8939  \\
        HAN-Daily(BERT) -  CTB-St & 0.5909 & 0.8422  \\
        HAN-Daily(BERT) -  CTB-Emo & 0.4772 & 0.7421  \\
        HAN-Daily(BERT) -  CTB-Tsa & 0.7359 &0.9114  \\
        \hline
        HAN-User - CTB & 0.7499 & 0.9194  \\ 
        HAN-User- CTB-St & 0.7396 & 0.9098  \\ 
        HAN-User- CTB-Emo & 0.7428 & 0.9108  \\ 
        HAN-User- CTB-Tsa & 0.7649 & 0.9198  \\ 
        \hline
        Mood\&Content - CTB-St & 0.5514 & 0.8298  \\ 
        Mood\&Content - CTB-Emo & 0.5364 & 0.8477  \\ 
        Mood\&Content - CTB-Tsa & 0.5129 & 0.8235  \\ 
        \hline
        Mood2Content - CTB-St & 0.8029 & 0.9313  \\ 
        Mood2Content - CTB-Emo & 0.7978 & 0.9299  \\ 
        Mood2Content - CTB-Tsa &\textbf{ 0.8116} & \textbf{0.9317}  \\ 
        \hline
    \end{tabular}
}
\label{Table: performance of ablation study}
\end{table}

\section{CASE STUDY}
Our framework's attention-based user encoder allows us to visualize the impact of daily tweets on the final depression prediction. This is accomplished by analyzing the assigned attention weight for each day. The daily attention weight of a positive case is depicted in Table \ref{Table: case study}, with a darker background color indicating a greater attention weight. On days 0, 2, and 16, the patient posted more desperate tweets, which are characterized by greater attention weights. This demonstrates that not only can our framework accurately predict depression risk but also estimate risk days. It is possible to delve deeper into the relationships between depression and social factors among a large number of depression patients by incorporating additional information such as weekdays vs. weekends or holidays vs. normal days. In addition, Table \ref{Table: case study} also lists the daily emotion of the patient, and we discovered that our model does not always emphasize negative emotions (such as sadness and anger). This suggests that the context-based model has a different emphasis than emotions, and the combination of both information sources can result in more accurate predictions. 
\begin{table}
  \caption{A case study of COVID-19 patient at depression risk. \\(Each daily tweet was performed emotion recognition by CTB-Emo and colored by its weight to user presentation.)}
  \resizebox{\linewidth}{!}{
  \begin{threeparttable}
  \begin{tabular}{ccc}
    \toprule
    \textbf{Days (Before)} & \textbf{Emotion} & \textbf{Daily Tweet}\\
    % \bottomrule
    \midrule
    Day 28 &Sadness & \makecell[c]{\colorbox{red!8.25}{\parbox{\columnwidth}{{\strut{I have had 3 telemed visits with my doctor. They are not really seeing people in person in my area...}}}}} \\
    \hline
    \multicolumn{3}{c}{......} \\
    \hline
    Day 17 &Optimism & \makecell[c]{\colorbox{red!4.88}{\parbox{\columnwidth}{{\strut{Ongoing mission to find new life and new civilizations. Boldly go where no one has gone before.}}}}} \\
    \hline
    Day 16 &Optimism & \makecell[c]{\colorbox{red!23.62}{\parbox{\columnwidth}{{\strut{I have covid, I have the antibodies. I was only very sick for 4 days. Then my immune system kicked in and i felt much better. it does not have to be a long battle. stay up as much as possible. don't take it laying down. don't sleep laying flat. eat and drink plenty. This is how i want to die.}}}}} \\
    \hline
    Day 15 &Optimism & \makecell[c]{\colorbox{red!4.78}{\parbox{\columnwidth}{{\strut{@user glad we talked about your problems and made it over that. each one teach one. ...}}}}} \\
    \hline
    \multicolumn{3}{c}{......} \\
    \hline  
    Day 8 &Sadness & \makecell[c]{\colorbox{red!22.88}{\parbox{\columnwidth}{{\strut{I developed a mental health disorder in which I crave ice cream. New song upload. ... }}}}} \\
    \hline
    Day 7 &Anger & \makecell[c]{\colorbox{red!6.38}{\parbox{\columnwidth}{{\strut{@user Yes, I believe only the best can apply for police, the only problem is that only \#offensive and \#offensive want this job now. }}}}} \\
    \hline
    \multicolumn{3}{c}{......} \\
    \hline  
    Day 1 &Null* & \makecell[c]{\colorbox{red!8.25}{\parbox{\columnwidth}{{\strut{This is not bf6. It is a demo of dice current tech, probably  taken from the last update of bf5. With less drug testing, less probation violations.}}}}} \\
    \hline
    Day 0 &Anger & \makecell[c]{\colorbox{red!30.00}{\parbox{\columnwidth}{{\strut{... I'm sure investigation will uncover solid evidence of a liberal conspiracy. It made me a mental case for a month after i had it. ...}}}}} \\
    \bottomrule
    \end{tabular}
    \begin{tablenotes}
    % \item This is note content.
    \item *Null indicates that the probability of all emotions is less than 0.5. \\ $\#$Offensive words have been replaced
    \end{tablenotes}
    \end{threeparttable}
    \label{Table: case study}
    }
\end{table}

% \parbox[position][height][inner-pos]{width}{text}

\section{ETHNIC CONSIDERATION}
For the protection of vulnerable individuals, privacy and ethical considerations are of paramount importance in the field of mental health. Using publicly accessible data collected via Twitter's official API, our study adheres to these stringent requirements. Our research utilized tweets obtained in accordance with Twitter's Privacy Policy, which informs users that the content they post on the platform, including their social profiles and tweets, is public and freely accessible to third parties. To protect individual privacy, we omitted usernames from our study and only provided the Tweet ID for download via the Tweet API. 

\section{CONCLUSION}
The COVID-19 pandemic has been three years, but the negative impact of COVID-19 still exists and tends to last for a long time. One critical social problem is the mental health risk of COVID-19 patients. COVID-19 triggers a non-trivial increase in depression patients. To alleviate this problem, one crucial step is to detect depressed COVID-19 patients as soon as possible and conduct an early intervention. This paper targets this critical social problem and the contributions of this paper are three folds: 1) We propose a novel research topic: predict the early depression risk with social media data; 2) We build a dataset from Twitter users which consists of 10,656 COVID-19 patients. 1,776 are positive cases who will emit depression signal after infection; 8,880 are in the control group who don't get depressed after infection. For each positive user in this dataset, we have the timestamp of COVID-19 infection and depression signal emergence as well as all posted tweets; 3) We also propose the Mood2Content model which manages to detect early depression risk. Mood2Content achieves an AUROC of 0.9317 in predicting the depression risk two weeks ahead of time, which outperforms baseline models ranging from popular machine learning models to pre-trained large language models. This enables the feasibility of early intervention of depressed patients. 

\section{LIMITATION}
Several potential limitations should be considered for this study. First, although we have taken numerous steps to identify eligible individuals as precisely as possible, it is possible that the dataset still contains some false positive cases. However, manual validation was performed to confirm the dataset's dependability, and the vast quantity of social media data helps to mitigate this issue. Second, we only encoded the first 256 tokens of daily tweets as sentence embeddings in order to meet the length limit of large language models and improve model efficiency. It may result in some information loss. Nonetheless, the threshold of 256 tokens covers 93\% of tweets, which mitigates the issue to some extent. Lastly, we did not use information about COVID-19 disease, such as symptoms, to enhance the model performance. We intend to investigate this in future studies.
% \begin{acks}
% To RoBERT, for the bagels and explaining CMYK and color spaces.
% \end{acks}

\bibliographystyle{ACM-Reference-Format}
\bibliography{reference}


\end{document}